\documentclass[a4paper,10pt,conference]{ieeeconf}

\IEEEoverridecommandlockouts
\usepackage{cite}
\usepackage{amsmath,amssymb,amsfonts}
\usepackage{graphicx}
\usepackage{textcomp}
\usepackage{xcolor}
\usepackage{booktabs}
\usepackage{array}
\usepackage{url}
\usepackage{stfloats}
\usepackage[hidelinks]{hyperref}

\graphicspath{{Figures/}}

\title{\LARGE \bf
Body-Motion Control of a Simulated Aerial Swarm from a First-Person View}

\author{
\authorblockN{
Yang Chen$^{\dagger}$,
Darius Giannoli$^{\dagger}$, and
Dario Floreano,
\IEEEmembership{Fellow, IEEE}
}
\thanks{$^{\dagger}$Y. Chen and D. Giannoli contributed equally to this work.}%
\thanks{All authors are with the Laboratory of Intelligent Systems,
\'{E}cole polytechnique f\'{e}d\'{e}rale de Lausanne (EPFL),
Lausanne, Switzerland (e-mail: chenyang@ai.iit.tsukuba.ac.jp;
darius.giannoli@epfl.ch; dario.floreano@epfl.ch).}%
\thanks{Videos are available in the
\protect\href{https://www.youtube.com/playlist?list=PLTn4QPxGFMEU}
{online video playlist}.}%
\thanks{This work was supported by the Swiss National Science Foundation
under Grant 200020\_212077.}%
\thanks{This study involving human participants was approved by the EPFL
Human Research Ethics Committee under Application No.\ HREC 092-2023.}%
\thanks{\copyright\ 2027 IEEE. Personal use of this material is permitted.
Permission from IEEE must be obtained for all other uses, in any current or
future media, including reprinting/republishing this material for advertising
or promotional purposes, creating new collective works, for resale or
redistribution to servers or lists, or reuse of any copyrighted component of
this work in other works.}%
}

\begin{document}

\maketitle
\thispagestyle{empty}
\pagestyle{empty}

\begin{abstract}

First-person-view (FPV) teleoperation of aerial swarms requires an operator to coordinate collective translation, viewing direction, and formation spacing. We present an upper-body interface that maps torso inclination, hand position, and head rotation to five continuous command dimensions. Neutral postures and motion ranges are calibrated for each participant. In a within-subject study, 14 participants navigated a simulated 15-agent swarm through three-dimensional obstacle courses using this interface and a conventional transmitter. Body-motion control reduced completion time by 19.4\% and centroid path length by 7.0\%, and increased path directness. Delivered-command variation was 88.8\% lower, and concurrent command changes were more frequent. These command measures characterize the complete interfaces, which differed in calibration and filtering. No differences were detected in gate-centering error, collection yield, crash or disconnection counts, overall workload, or usability. All participants reported higher physical demand with body-motion control. The implemented interface therefore improved FPV navigation efficiency at the cost of greater physical demand.

\end{abstract}

\section{Introduction}
Aerial swarms can extend the sensing range and coverage of a single
robot, making them useful for missions such as search and rescue.
Human judgment and oversight remain important in these
missions~\cite{kolling2016survey,hocraffer2017meta}. However, assigning
an operator to each robot does not scale. This motivates interfaces
that allow a single operator to command the swarm as a whole.

In three-dimensional environments, passing through an opening may require coordinated changes in planar translation, altitude, formation spacing, and viewing direction. An interface that supports these adjustments simultaneously may reduce the need to alternate between controls.

The viewpoint further constrains control. A top-down view (TDV) reveals the swarm's
spatial extent, connectivity, and relationship to nearby obstacles. However, it
requires an external viewing platform or real-time scene reconstruction. The
former can introduce a single point of failure, while the latter adds
computation and sensing requirements. A first-person view (FPV) can instead be
streamed from a \emph{focal agent}, the swarm member whose onboard camera
currently supplies the operator's view. The source can transfer if that agent
becomes unavailable~\cite{ben_focaldrone}. However, the limited field of view
and occlusions hide parts of the swarm and environment, while the viewpoint
moves, rotates, and may transfer between agents.

\begin{figure}[!t]
\centering
\includegraphics[width=0.9\columnwidth]{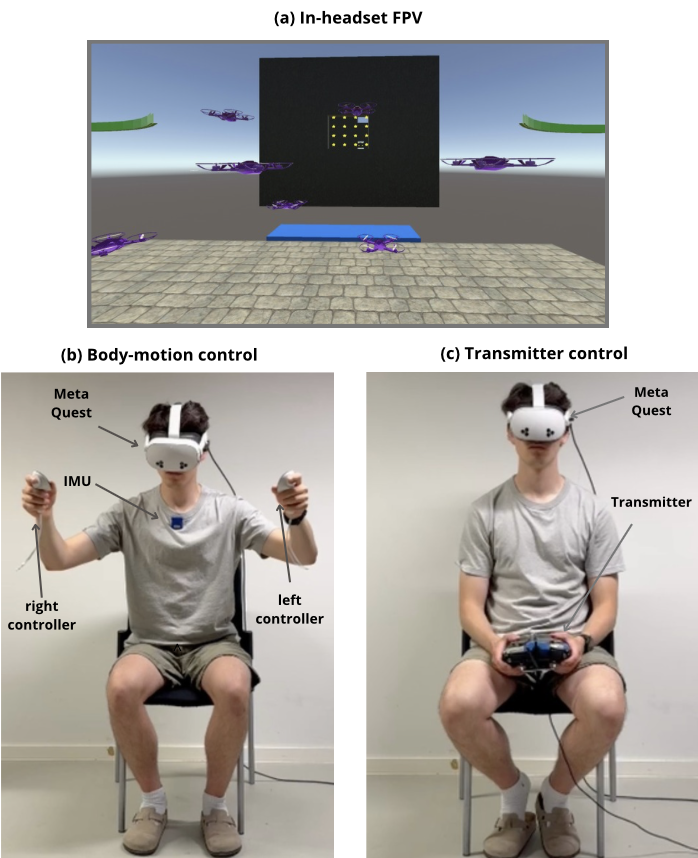}
\caption{Experimental interfaces and common visual feedback:
(a) the in-headset rear-focal FPV view, (b) body-motion control using
a torso-mounted IMU and hand controllers, and (c) transmitter control.
Both conditions used the same headset, visual feedback, and swarm
command space.}
\label{fig:study_overview}
\end{figure}

Jarvis et al.\ found that participants preferred views from the rear
of the swarm~\cite{ben_focaldrone}. Their commercial transmitter
mapped planar translation, vertical motion, view yaw, and formation
spacing to two sticks and a rotary knob. However, it remains unclear whether this layout is
well suited to the concurrent adjustment of five continuous DOF.
Jacob et al.\ showed that performance improves when the control structure of an input device matches the perceptual structure of the task~\cite{jacob1994integrality}, and input devices differ in how well they support coordinated control of multiple DOF~\cite{zhai1998quantifying}. On the transmitter, five commands are assigned to two thumbs and a knob, so changing spacing while translating and turning may require sequential rather than simultaneous inputs.

An upper-body interface can instead distribute spatially corresponding commands
across the head, torso, and hands. Body--machine interfaces have already enabled
immersive and personalized control of individual drones~\cite{rognon2018flyjacket,miehlbradt2018datadriven,
macchini2024datadriven}. Macchini et al.\ observed that more complex robots—differing in both morphology and DOF—were associated with the use of more body segments and greater inter-user variability in spontaneous control motions, strengthening the case for personalized mappings~\cite{macchini2024datadriven}. Together, these findings motivate participant-specific calibration for the present five-DOF task.

At swarm level, personalized hand
motion has supported four-dimensional control from an external viewpoint~\cite{macchini2021personalized}, while tactile and gesture-based systems have
supported other forms of swarm and formation control~\cite{tsykunov2019swarmtouch,kratky2025gesture}. Together, these studies support multidimensional body control but do not establish its effectiveness for a swarm viewed through a moving focal agent. 

We therefore investigate whether distributing five continuous swarm commands
across the head, torso, and hands can address the input bottleneck of
focal-agent FPV swarm teleoperation. We implement a participant-calibrated, upper-body interface for controlling a simulated drone swarm and compare it with a conventional transmitter in a within-subject study of 14 participants navigating three-dimensional obstacle courses. The contributions are (i) an upper-body mapping that distributes five swarm commands across the
torso, hands, and head, aligned with the focal-agent reference frame during
view rotation and transfer, and (ii) a within-subject evaluation of navigation
efficiency, command behavior, task outcomes, and user experience.

\section{Swarm Control System and Input Interfaces}

\subsection{Shared Swarm and FPV System}
\label{sec:system}
Both input conditions generated a common five-dimensional command vector for the simulated 15-agent swarm:
\begin{equation}
\mathbf{u}_t =
\left[
v_{\mathrm{fwd},t}^{\mathrm{ref}},
v_{\mathrm{lat},t}^{\mathrm{ref}},
v_{\mathrm{vert},t}^{\mathrm{ref}},
\dot{\psi}_{t}^{\mathrm{cmd}},
\rho_t^{*}
\right]^{\mathsf T}.
\label{eq:swarm_command}
\end{equation}
The first three components specify focal-frame horizontal and world-frame vertical velocities; the remaining components specify FPV yaw rate and target agent spacing.

A decentralized Olfati--Saber controller~\cite{olfatisaber2006flocking}
combined spacing regulation, velocity alignment, reference-velocity tracking,
and obstacle repulsion. Surviving agents formed an undirected interaction
graph, with an edge between agents whose Euclidean separation was below
\(R_t=\max(2\rho_t^*,\,\rho_t^*+2~\mathrm{m}).\)
The largest connected component defined the connected swarm; surviving agents
outside it were classified as disconnected. The simulation built on the FPV
swarm-teleoperation framework of Palle et al.~\cite{palle2026multisensory}.

Both conditions used the same stereoscopic rear-focal FPV.
The rear viewpoint was motivated by the performance and
preference results reported by Jarvis et al.~\cite{ben_focaldrone}.
The focal agent was selected from the rear of the swarm along
the horizontal viewing direction. The feed transferred to another
agent only when it remained at least 0.5~m behind the current
focal agent along this direction for at least 0.5~s.

In both conditions, participants viewed the FPV through a virtual reality (VR)
headset (Meta Quest 3S, Meta, USA). For body-motion control, the headset measured
head yaw, two handheld controllers measured hand positions, and a chest-mounted
inertial measurement unit (LPMS-B2, LP-Research, Japan) measured torso
inclination. The swarm controller, command limits, focal-agent logic, and
simulation were identical across conditions. The interfaces
differed in input hardware, command mapping, participant-specific
calibration, and temporal filtering.

\begin{figure*}[t]
\centering
\includegraphics[width=\textwidth]{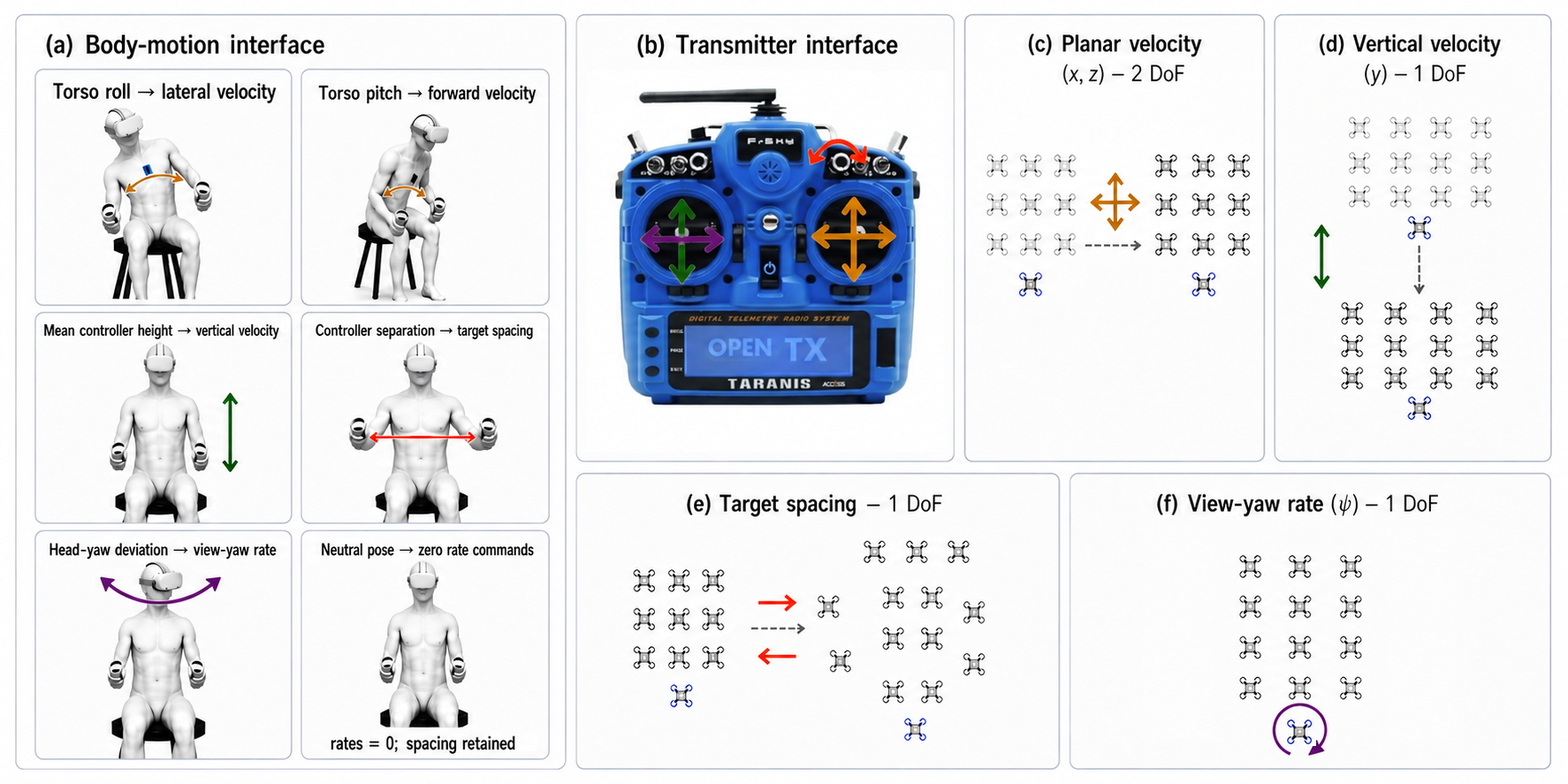}
\caption{Control mappings for the body-motion and transmitter interfaces:
(a) body-to-swarm mappings, (b) transmitter mappings, and (c)--(f) the
resulting swarm translation, spacing, and view rotation.}
\label{fig:mapping}
\end{figure*}

\subsection{Body-to-Swarm Mapping}
\label{sec:mapping} 
The body-to-swarm mapping linked body movements to spatially corresponding
swarm commands and allowed several commands to be adjusted
simultaneously. Torso inclination has previously been used to command the
pitch and roll of a single fixed-wing drone flying at constant
speed~\cite{rognon2018flyjacket}. Here, forward, backward, and
lateral leaning instead commanded the planar velocity of the swarm.
The hands and head commanded the remaining three DOF:
raising or lowering both hands commanded vertical motion,
hand separation set target agent spacing, and head rotation
commanded FPV yaw. Participant-specific calibration accounted
for differences in posture and range of motion.

At time $t$, the interface measured sagittal and lateral torso
inclinations $(\theta_t,\phi_t)$, headset yaw $\psi_t$, mean
controller height
\[
h_t=\frac{y^W_{L,t}+y^W_{R,t}}{2},
\]
and controller separation
\[
s_t=\left\lVert\mathbf{p}^W_{R,t}-\mathbf{p}^W_{L,t}\right\rVert.
\]
Here, superscript $W$ denotes the world frame, and subscripts
$L$ and $R$ identify the left and right controllers.
The vector $\mathbf{p}$ denotes controller position, and $y$
is its vertical coordinate. Torso inclinations were measured
relative to the calibrated neutral posture.
Translation and yaw used rate control, so a sustained posture produced
continuous motion; hand separation directly set target agent spacing.

\emph{Participant-specific normalization:}
Each rate-controlled signal was mapped to $[-1,1]$ using its calibrated lower
extreme, neutral value, and upper extreme $(q^{-},q^{0},q^{+})$, with a
deadzone of half-width $d$ around neutral. Let
$q_{\mathrm{L}}=q^{0}-d$ and $q_{\mathrm{U}}=q^{0}+d$. The normalization was

\begin{equation}
\begin{split}
&\mathcal{D}(q_t;q^{-},q^{0},q^{+},d)\\
&\qquad=
\begin{cases}
\max\!\left(-1,\dfrac{q_t-q_{\mathrm{L}}}{q_{\mathrm{L}}-q^{-}}\right),
& q_t<q_{\mathrm{L}},\\[8pt]
0,
& q_{\mathrm{L}}\leq q_t\leq q_{\mathrm{U}},\\[8pt]
\min\!\left(1,\dfrac{q_t-q_{\mathrm{U}}}{q^{+}-q_{\mathrm{U}}}\right),
& q_t>q_{\mathrm{U}}.
\end{cases}
\end{split}
\label{eq:deadzone_mapping}
\end{equation}
Separate scaling on either side of neutral accommodated asymmetric ranges of
motion.

\emph{Translation:}
The normalized translation commands were
\begin{align}
c_{\mathrm{fwd},t}
&=\mathcal{D}(\theta_t;\theta^{-},0,\theta^{+},5^{\circ}),
\nonumber\\
c_{\mathrm{lat},t}
&=\mathcal{D}(\phi_t;\phi^{-},0,\phi^{+},5^{\circ}),
\nonumber\\
c_{\mathrm{vert},t}
&=\mathcal{D}(h_t;h_{\min},h_0,h_{\max},0.05~\mathrm{m}).
\label{eq:body_rate_commands}
\end{align}
Positive values commanded forward, rightward, and upward motion. After
smoothing, the normalized commands were multiplied by
$10~\mathrm{m\,s^{-1}}$ to obtain the reference velocities in
Eq.~\eqref{eq:swarm_command}.

The forward and lateral velocities were defined relative to the displayed
heading and updated after FPV rotation or focal-agent transfer, so that leaning
forward always moved the swarm toward the center of the view.

\emph{Inter-agent spacing:}
Controller separation was mapped linearly from $[s_{\min},s_{\max}]$ to target
agent spacing $\rho_t\in[1,5]~\mathrm{m}$, then clipped and smoothed to produce
the applied command $\rho_t^*$.

\emph{FPV yaw:}
With
$\Delta\psi_t=
\operatorname{wrap}_{[-180^{\circ},180^{\circ})}
(\psi_t-\psi_{0})$
denoting the head-yaw deviation from its neutral orientation, the FPV yaw-rate
command was
\begin{equation}
\dot{\psi}^{\,\mathrm{cmd}}_t
=
48^{\circ}\mathrm{s}^{-1}
\mathcal{D}\!\left(
\Delta\psi_t;\psi_{\mathrm{L}},0,\psi_{\mathrm{R}},10^{\circ}
\right).
\label{eq:yaw_rate}
\end{equation}
Returning the head to neutral stopped FPV rotation without restoring the
previous viewing direction.

Pilot testing was used to tune controller parameters and refine the
body-to-swarm mapping.

\subsection{Participant-Specific Calibration}
\label{sec:calibration}
Each participant completed a one-time guided 13-pose calibration to record
seated neutral postures and comfortable motion limits. The chair configuration and Quest tracking origin remained fixed thereafter.

Calibration included five torso poses (neutral and maximum forward, backward,
left, and right), three head-yaw poses (neutral and maximum left and right),
two controller-separation poses (minimum and maximum), and three
controller-height poses (minimum, neutral, and maximum). Neutral torso and head poses defined the IMU orientation offset and headset yaw
reference $\psi_0$, respectively; the controller-separation limits defined the
agent-spacing range.

Table~\ref{tab:calibration_summary} summarizes the calibrated spans. Angular
spans combine opposing excursions, while position spans are
maximum--minimum differences.

\subsection{Transmitter Baseline}
\label{sec:transmitter}
A conventional drone transmitter (Taranis X9D Plus, FrSky, China) connected to
the PC by USB generated the same command vector $\mathbf{u}_t$. The right stick
controlled planar velocity, the left horizontal axis controlled FPV yaw rate,
the non-self-centering throttle controlled vertical velocity, and the right
rotary knob set target agent spacing. Stick axes used symmetric deadzones, and
the knob mapped linearly to $\rho^*\in[1,5]$~m. Pilot-tested gains and deadzones were fixed across participants. Unlike the body-motion mapping, the off-the-shelf transmitter used standardized controls and received no participant-specific calibration or additional temporal
smoothing.

\begin{table}[t]
\centering
\caption{Calibrated input spans across participants ($N=14$).}
\label{tab:calibration_summary}
\footnotesize
\setlength{\tabcolsep}{3.5pt}
\begin{tabular}{@{}lcc@{}}
\toprule
Input span & Mean $\pm$ SD & Range \\
\midrule
Sagittal torso ($^\circ$)       & $56.4 \pm 18.4$  & $34.2$--$98.1$ \\
Lateral torso ($^\circ$)        & $52.0 \pm 18.0$  & $26.2$--$81.2$ \\
Head yaw ($^\circ$)             & $117.5 \pm 24.7$ & $65.6$--$138.0$ \\
Controller separation (m)       & $1.10 \pm 0.33$  & $0.70$--$1.51$ \\
Controller height (m)           & $0.77 \pm 0.22$  & $0.43$--$1.22$ \\
\bottomrule
\end{tabular}
\end{table}

\section{Experimental Evaluation}
\label{sec:experiment}

\subsection{Participants and Study Design}
\label{sec:participants_design}

The study included $N=14$ adults recruited from the local university
community (10 men and 4 women; age: $22.4 \pm 4.0$ years, range 19--35).
Participants self-reported being able to perform the seated upper-body
movements and use the VR headset. All
participants completed the study, provided written informed consent, and were
compensated.

We used a two-condition within-subject design. The first participant's starting
interface was randomized and alternated thereafter, yielding seven participants
per sequence. Sessions lasted approximately 1~h and were completed while
seated.

\subsection{Task and Procedure}
\label{sec:task}
All participants viewed the same prerecorded instructions and demonstrations.
They were instructed to complete each course quickly, pass near gate centers,
collect stars, and minimize crashes and disconnections. They viewed only the
stereoscopic FPV, with no timer, score, or task-status display.

Each recorded course contained seven gates whose positions, heights, and
opening sizes were varied to require horizontal translation, altitude
adjustment, and formation-spacing control. Each course included one 90-degree turn, left in one course and right in the other, requiring participants to rotate the FPV and continue in the rotated focal-agent reference frame. Stars distributed across the gate openings encouraged control of swarm extent rather than only the focal agent or centroid; collection yield measured this collective performance.

A star was collected when any agent entered its trigger. Wall or inter-agent
collisions removed the affected agent, and surviving agents outside the largest
connected component were classified as disconnected at each sample.

Before the body-motion block, participants completed the calibration described
in Section~\ref{sec:calibration}. Each interface block began with familiarization
on the practice course: participants first maneuvered freely with standardized
experimenter guidance and then completed one unrecorded full-course traversal.
Two recorded runs followed in fixed order.

After each block, participants completed the weighted NASA-TLX \cite{hart1988nasatlx} and SUS~\cite{brooke1996sus} for the two recorded runs.
After both blocks, they reported interface preference, relative motion sickness,
and optional comments.

\subsection{Evaluation Metrics}
Completion time was the primary outcome, consistent with task-oriented HRI
evaluation~\cite{steinfeld2006metrics}. Secondary measures covered trajectory characteristics, task performance, swarm integrity, delivered commands, and subjective experience.

\emph{Traversal efficiency.}
Completion time $T$ extended from the recorded start event to the finish event.
Centroid path length $L$ was the total three-dimensional distance traveled by
the controlled-component centroid, and mean speed was $\bar v=L/T$. Path length
and speed were treated as supporting outcomes.

\emph{Trajectory characteristics.}
Path directness was the sum of the straight-line endpoint distances across
eight course segments, divided by the total path length $L$
\cite{mavrogiannis2022socialmomentum}:
\begin{equation}
D_{\mathrm{path}}
=
\frac{1}{L}
\sum_{j=1}^{8}
\left\|
\mathbf{x}_{j,\mathrm{end}}
-
\mathbf{x}_{j,\mathrm{start}}
\right\|_2.
\label{eq:path_directness}
\end{equation}
The eight segments ran from the start to the first gate, between consecutive
gates, and from the final gate to the finish.
\begin{figure}[t]
\centering
\includegraphics[width=0.9\columnwidth]{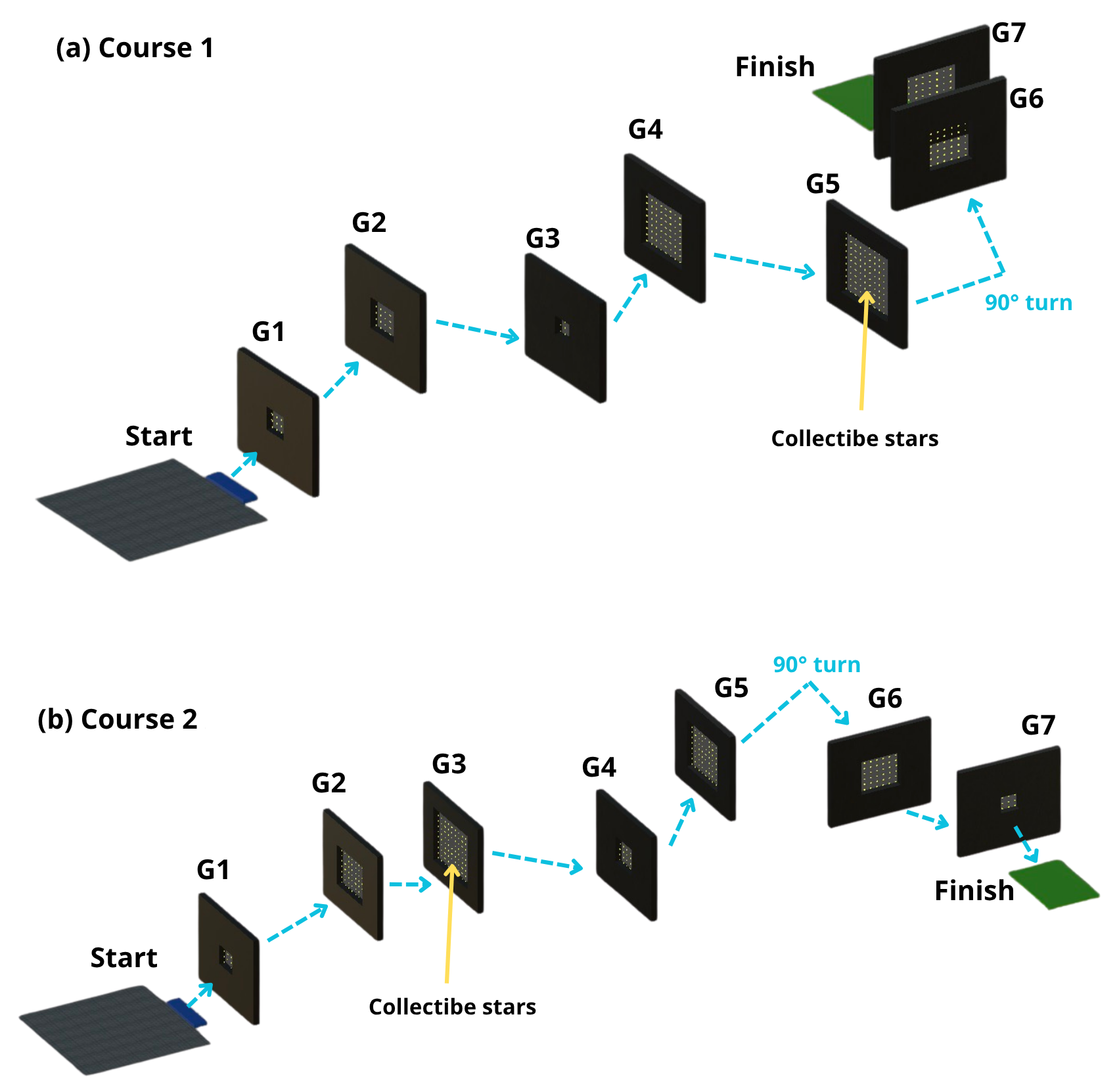}
\caption{Obstacle course layouts.
Labels G1--G7 mark the gate sequence; dashed arrows show traversal direction.}
\label{fig:course_layouts}
\end{figure}

\begin{figure*}[t]
\centering
\includegraphics[width=0.9\textwidth]{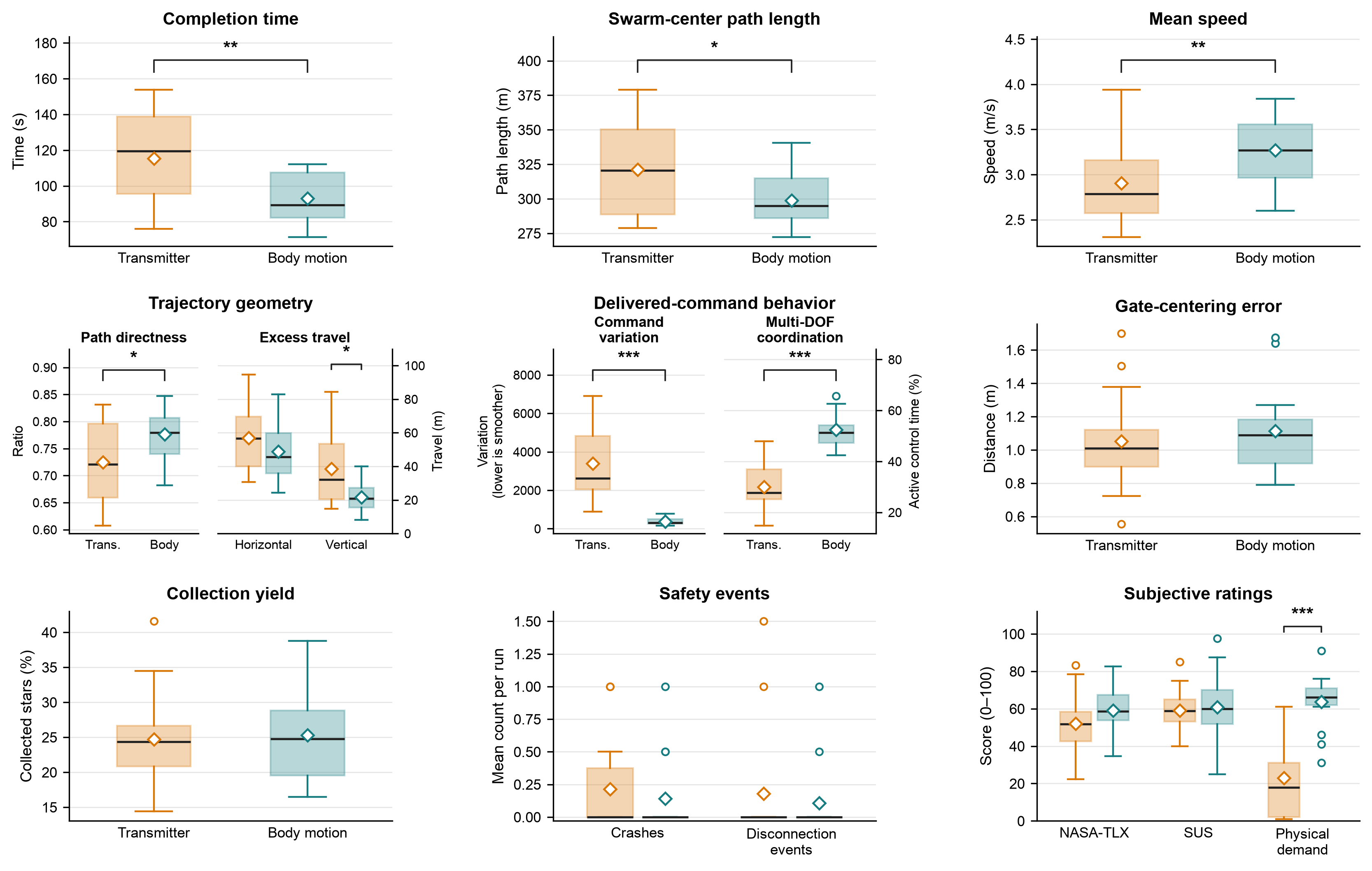}
\caption{Participant-level outcomes for transmitter and body-motion control
($n=14$). 
Boxes show interquartile ranges with median lines; diamonds indicate means
and circles indicate outliers. Orange denotes transmitter and teal denotes body
motion. Stars mark Holm-adjusted paired comparisons
($*p<0.05$, $**p<0.01$, $***p<0.001$).}
\label{fig:paired_results}
\end{figure*}

For each segment, excess horizontal travel $H_{\mathrm{exc}}$
was the horizontal path length in the $x$--$z$ plane minus the
straight-line horizontal distance between its endpoints.
Excess vertical travel $V_{\mathrm{exc}}$ was the total absolute
height change minus the absolute difference between the segment's
start and end heights. Both measures were summed across the eight
course segments. Lower values indicate less excess motion.

\emph{Gate-centering error.}
For each gate, the swarm centroid was linearly interpolated at the gate-plane
crossing. Error was the in-plane distance $\sqrt{d_h^2+d_v^2}$ from the gate
center, where $d_h$ and $d_v$ are the horizontal and vertical offsets. We
averaged this distance across the seven gates.

Collection yield was the fraction of the 316 available stars collected.
Crash count was the number of agents that crashed during a run.
Disconnection count was the number of agent-disconnection events recorded
during a run.

\emph{Delivered-command behavior.}
Each command was normalized to $[0,1]$. Let $\mathbf{q}_k$ be the five-command
vector at sample $k$ and $\tau=t_M-t_1$ the run duration. Delivered-command
variation was
\begin{equation}
E_{\mathrm{cmd}} = \frac{\tau}{5} \sum_{k=1}^{M-1} \frac{\lVert\mathbf{q}_{k+1}-\mathbf{q}_k\rVert_2^2} {t_{k+1}-t_k}.
\label{eq:command_metrics}
\end{equation}
Lower values indicate smoother delivered commands.

\emph{Multi-DOF coordination.}
A command was considered changing when the absolute rate of
change of its normalized value was at least $0.10\,\mathrm{s^{-1}}$.
Active control time comprised intervals with at least one
changing command. Multi-DOF coordination was the percentage
of active control time with at least two changing commands.
Heatmap diagonal entries show individual-command activity, including
concurrent changes in other commands; off-diagonal entries show
pairwise co-change, including intervals when additional commands
were changing. These percentages overlap and are not additive.
We also partitioned active control time by the number of changing
commands (1--5), yielding mutually exclusive percentages that sum
to 100\%.
Because the interfaces used different temporal filtering, these
measures describe the complete delivered-command pipelines rather
than intrinsic smoothness or intentional coordination.

\section{Results}
All 56 runs were retained; the two runs per interface were averaged for each
participant, yielding 14 paired observations. Interfaces were compared using exact two-sided Wilcoxon signed-rank tests ($\alpha=0.05$)~\cite{wilcoxon1945ranking}, with Holm adjustment within outcome families~\cite{holm1979procedure}.

\begin{figure}[!t]
\centering
\includegraphics[width=0.9\columnwidth]
{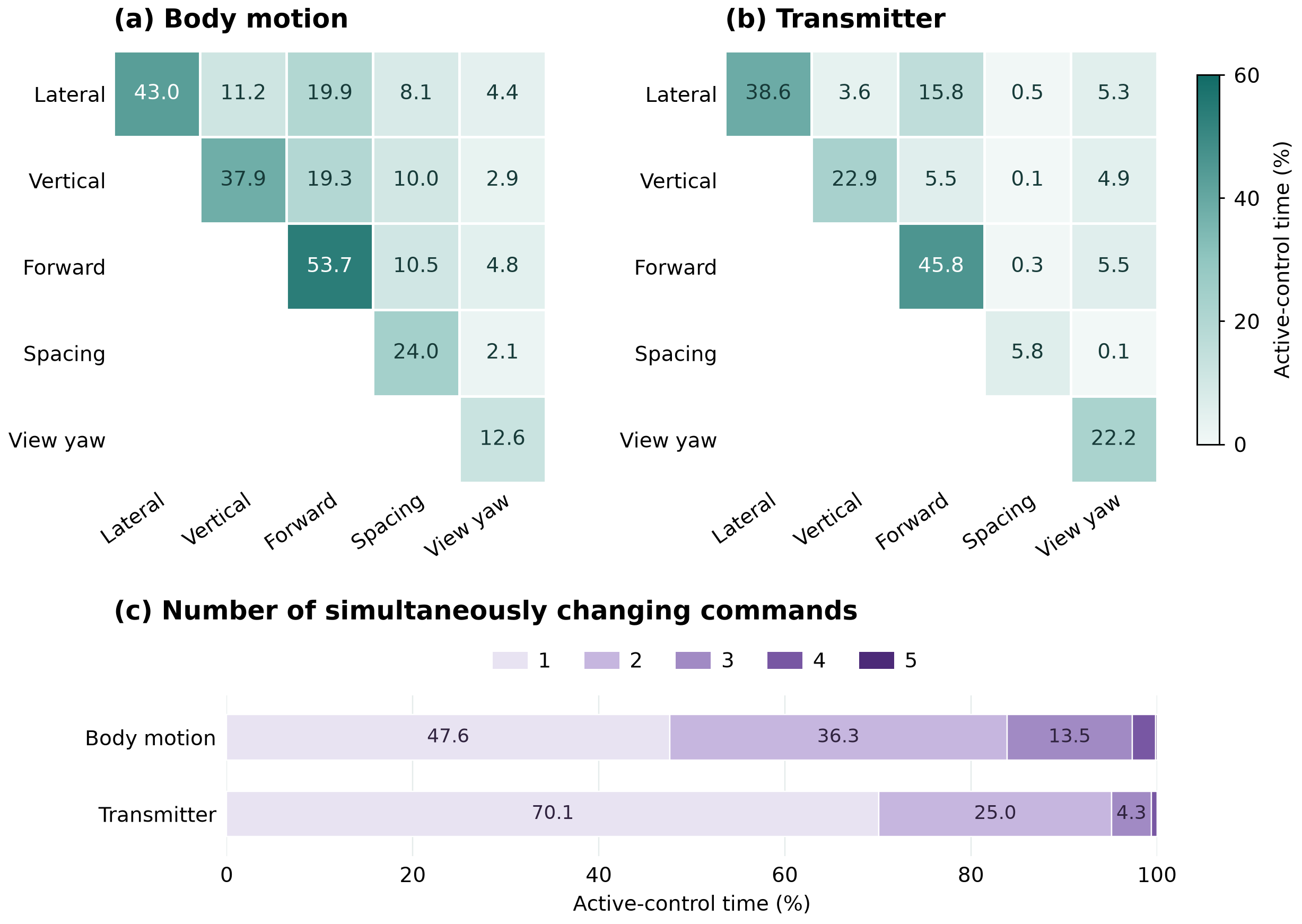}
\caption{Command activity for (a) body-motion and (b) transmitter
control. Heatmaps show individual-command activity (diagonal) and
pairwise co-change (upper triangle), including changes alongside
other commands. Percentages overlap and use a shared color scale.
(c) Time with exactly 1--5 changing commands; each bar totals 100\%.
Active-control time includes intervals when any command's absolute
normalized rate of change is at least $0.10\,\mathrm{s^{-1}}$.
Values are participant means ($n=14$), after averaging runs
within participants.}
\label{fig:dof_pairwise}
\end{figure}

\emph{Traversal efficiency.}
Completion time was 19.4\% lower with body motion, a mean reduction of
22.4~s (Holm-adjusted $p=0.001$); 13 of 14 participants were faster. Centroid
path length was 7.0\% shorter ($p=0.011$), and mean speed was 12.5\% higher
($p=0.001$).

\emph{Trajectory characteristics.}
With body-motion control, path directness increased by 0.052 (Holm-adjusted $p=0.032$), while excess vertical
travel decreased by 44.0\%, corresponding to 16.9~m (Holm-adjusted $p=0.049$).
Excess horizontal travel was 14.2\% lower with body motion (48.8 versus
56.8~m), but no clear difference was detected (mean difference $-8.0$~m,
$p=0.153$). Figure~\ref{fig:trajectory_overlays} illustrates these patterns.

\emph{Delivered-command behavior.}
With body-motion control, command variation was 88.8\% lower (Holm-adjusted $p<0.001$) and multi-DOF coordination was 22.5 percentage points higher than with transmitter control ($p<0.001$); every participant showed effects in the same direction. Exactly two commands changed during 36.3\% of active control time
with body motion versus 25.0\% with the transmitter; three or more
changed during 16.1\% versus 4.9\%
(Fig.~\ref{fig:dof_pairwise}(c)).
Pairwise co-change was higher with body motion for seven of ten
command pairs (Holm-adjusted $p\leq0.043$).
The largest increases were vertical--forward (19.3\% versus 5.5\%),
forward--spacing (10.5\% versus 0.3\%), and vertical--spacing
(10.0\% versus 0.1\%; $p=0.0012$ each).
Lateral--spacing co-change was also higher with body motion
(8.1\% versus 0.5\%; $p=0.0012$).

\emph{Task performance and swarm integrity.} 
No differences were detected in gate-centering error, collection yield, crash count, or disconnection-event count (Holm-adjusted \(p=1.000\) for all four outcomes). 

\emph{Subjective experience.} 
No differences were detected in overall weighted NASA-TLX or SUS
scores (mean differences, body motion minus transmitter: 7.2 and
1.6 points; Holm-adjusted $p=0.482$ and $p=0.491$, respectively).
However, all 14 participants reported higher physical demand with
body motion (mean increase 40.7 points; Holm-adjusted $p<0.001$).
Eight participants preferred body motion, five preferred the
transmitter, and one expressed no preference. Six reported greater
motion sickness with body motion, four with the transmitter,
and four reported no difference.

\begin{figure*}[!t]
\centering
\includegraphics[width=0.8\textwidth]
{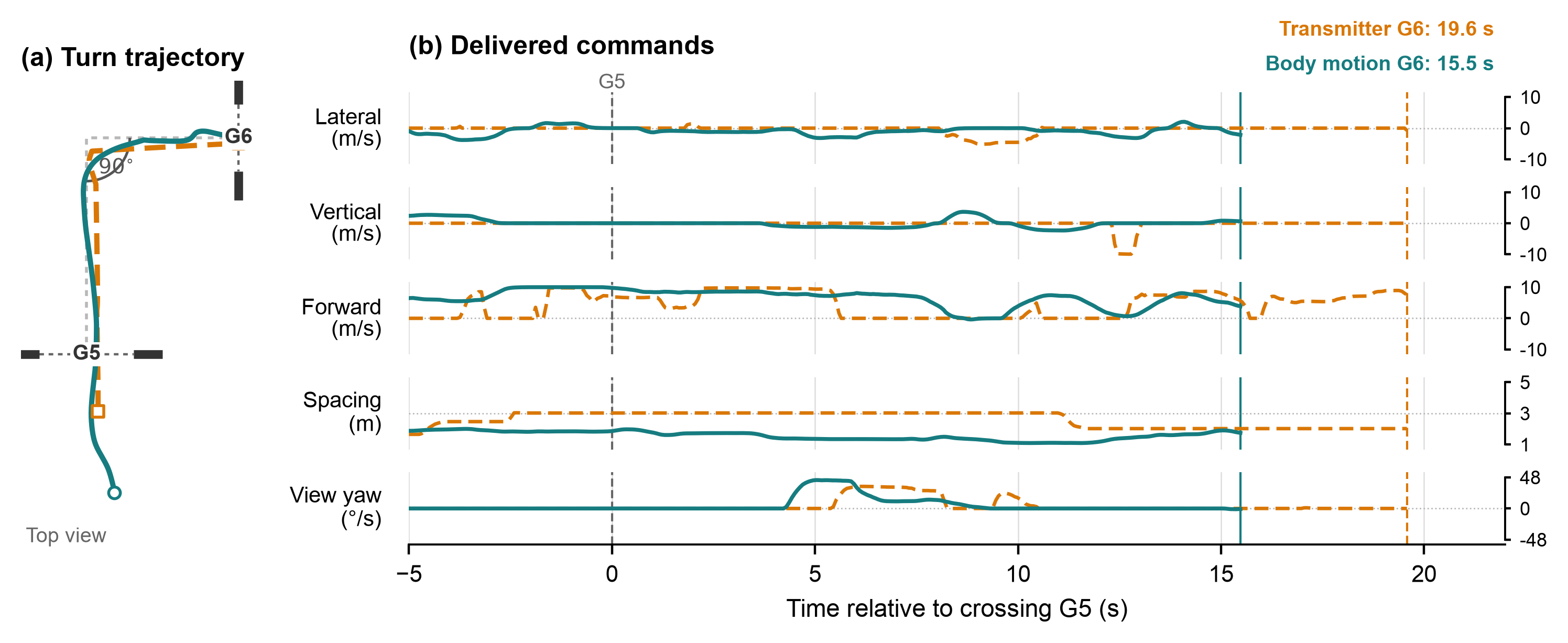}
\caption{Representative 90$^\circ$ turn, shown from 5~s before G5 to the G6 crossing. (a) Top-view swarm-centroid trajectories. (b) Delivered commands in physical units, aligned to G5 ($t=0$); solid teal indicates body motion and dashed orange indicates the transmitter. Vertical lines mark the G6 crossings. The delivered forward command was positive for 95.5\% of the displayed body-motion interval and 61.3\% of the displayed transmitter interval.}
\label{fig:traj_gate}
\end{figure*}

\section{Discussion}
The implemented body-motion interface improved traversal efficiency and path
directness in the tested FPV swarm-navigation task. Higher multi-DOF
coordination may partly explain these gains: participants reported adjusting
several commands together rather than moving between transmitter controls.
This is consistent with the benefit of matching the control structure of the
device to the task~\cite{jacob1994integrality}. Lower delivered-command
variation may also have supported more continuous motion and fewer corrections.
These associations do not establish causation.
\begin{figure}[!t]
\centering
\includegraphics[width=0.8\columnwidth]
{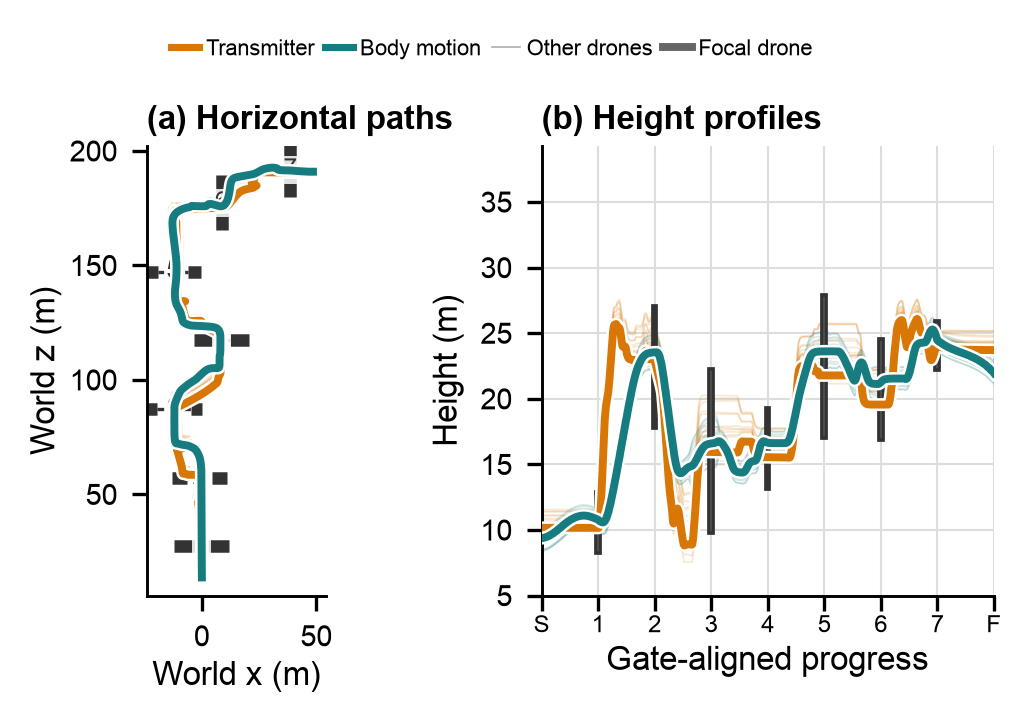}
\caption{Example Course 2 trajectories for one participant: (a) horizontal
paths and (b) gate-aligned heights under each interface. Thin lines show
non-focal agents, thick lines show the focal agent, and black markers show gate
locations and vertical openings.}
\label{fig:trajectory_overlays}
\end{figure}

Body-motion control was associated with less excess vertical travel.
The example in Fig.~\ref{fig:trajectory_overlays}(b) shows larger
vertical reversals under transmitter control, consistent with
repeated altitude corrections. However, this example does not
identify their cause. The body-motion altitude command was
individually calibrated and smoothed, whereas the transmitter
used a fixed gain and no additional smoothing. These results
therefore characterize the complete interfaces tested and do
not isolate the effects of input modality, calibration, or filtering.

Several participants reported that leaning forward also lowered their hands,
unintentionally changing the altitude command and requiring correction,
although they adapted quickly. This mechanical coupling could increase measured
multi-DOF coordination without reflecting intentional concurrent control.
Consistent with this concern, the largest increase in pairwise
co-change involved vertical and forward commands.
However, task demands and interface-specific filtering could also
produce this pattern, so the heatmaps do not identify its cause. Arm
support~\cite{rognon2018flyjacket} or measuring hand height relative to the
torso could help decouple these commands.

The efficiency gain was not accompanied by a detected loss of task accuracy or swarm integrity. Gate-centering error, collection yield, crash count, and disconnection events did not differ between interfaces. However, the absence of a detected difference does not establish equivalence. 

Although overall workload did not differ, all 14 participants rated body motion
as more physically demanding. Lower ratings on other TLX dimensions may have
offset this increase in the overall workload score. Preferences and relative
motion-sickness responses varied, indicating that the efficiency gain
was not accompanied by a uniform subjective advantage. Future designs could reduce the required arm movement by
increasing the hand-height gain, although higher gains may reduce fine control.

The main practical benefit is not access to additional commands, because both
interfaces controlled the same five DOF, but the ability to combine them in
time, as illustrated in Fig.~\ref{fig:traj_gate}. Together with the group-level
coordination result, this suggests that concurrent control may help operators resize and redirect the swarm without repeatedly interrupting the forward command. All participants showed higher measured coordination after brief familiarization.

These findings extend upper-body control from individual aerial
robots~\cite{rognon2018flyjacket,miehlbradt2018datadriven,
macchini2024datadriven} and externally viewed
formations~\cite{macchini2021personalized,tsykunov2019swarmtouch,
kratky2025gesture} to continuous five-DOF swarm control from a rear-focal FPV,
and complement transmitter-based FPV swarm
control~\cite{ben_focaldrone}. Generalization to other
viewpoints and physical swarms with communication, aerodynamic, and safety
constraints remains to be tested.

\section{Conclusion}
We presented a participant-calibrated upper-body interface for continuous
five-DOF FPV swarm teleoperation. In a 14-participant within-subject study,
the implemented body-motion control system reduced completion time by 19.4\% relative to a transmitter and produced more direct trajectories with lower delivered-command variation, at the
cost of higher physical demand and with no detected differences in task
accuracy, swarm integrity, total workload, or usability. Validation on physical
swarms remains necessary. For first-person-view swarm teleoperation, these results support
further evaluation of interfaces that distribute command dimensions
across body segments. Future designs should address physical demand
and unintended coupling between commands.

\section*{Acknowledgment}
The authors thank Benjamin Jarvis for constructive feedback on the manuscript, and Gabriel Taieb for helping test the experimental protocol.

\bibliographystyle{IEEEtran}
\bibliography{references}

\end{document}